\documentclass{article}

\usepackage[nonatbib, final]{neurips_2021}

\usepackage[utf8]{inputenc} 
\usepackage[T1]{fontenc}    
\usepackage{hyperref}       
\usepackage{url}            
\usepackage{booktabs}       
\usepackage{amsfonts}       
\usepackage{nicefrac}       
\usepackage{microtype}      
\usepackage{xcolor}         
\usepackage{graphicx}
\usepackage{amsmath}

\title{Maximum Entropy Model-based Reinforcement Learning}

\author{%
  Oleg Svidchenko \\
  JetBrains Research\\
  HSE University\\
  Saint Petersburg, Russia \\
  \texttt{oleg.svidchenko@jetbrains.com} \\
  \And
  Aleksei Shpilman \\
  JetBrains Research\\
  HSE University\\
  Saint Petersburg, Russia \\
  \texttt{aleksei@shpilman.com} \\
}

\newcommand{\BibTeX}{\rm B\kern-.05em{\sc i\kern-.025em b}\kern-.08em\TeX}

\begin{document}

\maketitle 

\begin{abstract}

Recent advances in reinforcement learning have demonstrated its ability to solve hard agent-environment interaction tasks on a super-human level. However, the application of reinforcement learning methods to a practical and real-world tasks is currently limited due to most RL state-of-art algorithms' sample inefficiency, i.e., the need for a vast number of training episodes. For example, OpenAI Five algorithm that has beaten human players in Dota 2 has trained for thousands of years of game time. Several approaches exist that tackle the issue of sample inefficiency, that either offer a more efficient usage of already gathered experience or aim to gain a more relevant and diverse experience via a better exploration of an environment. However, to our knowledge, no such approach exist for model-based algorithms, that showed their high sample efficiency in solving hard control tasks with high-dimensional state space. This work connects exploration techniques and model-based reinforcement learning. We have designed a novel exploration method that takes into account features of the model-based approach. We also demonstrate through experiments that our method significantly improves the performance of model-based algorithm Dreamer. 

\end{abstract}



\section{Introduction}
In recent years deep reinforcement learning have demonstrated the ability to solve complex control tasks such as playing strategic games with super-human performance~\cite{alphastar, openaifive}, controlling complex robotic systems~\cite{openairubicscube}, and others. However, most of popular reinforcement learning algorithms require a vast amount of environment interaction experience to achieve this result. For example, OpenAI~Five~\cite{openaifive} takes more than ten thousand years of simulated game time to master Dota 2. This limitation prevents reinforcement learning from being widely applied to real-world tasks.

Several methods attempt to solve the problem of sample efficiency. Some approach this problem as the problem of efficient use of already collected data and utilize the idea of experience prioritization \cite{prioritizedbuffer} or data augmentation and normalization to prevent overfitting to a small data sample \cite{curl, drqv2, spectralnormalization}. Others encourage agents to explore the environment more efficiently and collect more diverse experiences \cite{maxentexploration, curiosity, neurips2020exploration1, neurips2020exploration2}.

Model-based reinforcement learning algorithms create a representation of the environments to train an agent, rather than only interact with the original environment. This also leads to a significantly more sample efficient solutions than model-free approach~\cite{dreamer, dreamer2}. Recently, some researchers have used~\cite{curiosity,neurips2020exploration2} simple learned world models to assign intrinsic rewards for an agent. However, to our knowledge, no research exist, that bind efficient exploration methods with model-based algorithms. 

The main contribution of this work is that we create a connection between exploration methods and model-based reinforcement learning. We present a method of maximum-entropy exploration for model-based reinforcement learning agents and build a MaxEnt Dreamer that improves the performance to model-based Dreamer~\cite{dreamer}. We also introduce some additional modifications to improve the stability and perform an ablations study to show their effectiveness. Experimental results show that our exploration method and our modifications significantly improve the performance of Dreamer. We also perform an ablation study to show that it is indeed exploration techniques that contribute the majority of the improvement.

\section{Background}
\subsection{Reinforcement Learning}
Reinforcement Learning is a part of Machine Learning that focuses on solving agent-environment interaction tasks. In such tasks, the environment usually has state $s$ at each time moment, and the agent has an observation $o$ that somehow correlates a state $s$. At each time step, agent commits an action $a$ that leads to transition from state $s$ to some state $s'$. Then, agent receive a reward $r$ for transition $(s, a, s')$ and observation $o'$ that corresponds to a state $s'$.

More formally, in our work, we assume partially observable Markov decision process (POMDP) which consists of state space $S$, action space $A$, observation space $O$, transition function $T: S \times A \rightarrow S$, reward function $R: S \times A \times S \rightarrow \mathbb{R}$, observation function $f_O: S \rightarrow O$ and episode termination function $D: S \rightarrow \{0, 1\}$. All this functions may be stochastic. The goal is to find such a policy $\pi: O \rightarrow A$ that maximizes expected cumulative discounted reward $\mathbb{E}_{\tau | \pi} \sum_{i = 0}^T \gamma^i r_i$ where $\tau$ is a trajectory that consists of transitions $(o, s, a, s', o', r, d)$ acquired from an POMDP with respect to policy $\pi$, $T$ is a number of transitions in the trajectory $\tau$ and $\gamma \in (0, 1)$ is a discount factor. To estimate an effectiveness of a specific policy when start acting in a state $s$ there are a value function which is defined as:

\begin{equation} \label{eq: value-function}
    V_\pi(s) = \mathbb{E}_{a \sim \pi(.|f_O(s)),\ s' \sim T(.|s, a)} [r + (1-d) \cdot \gamma \cdot V_{\pi}(s'))]
\end{equation}
where $d = D(s, a, s')$ and $r = R(s, a, s')$. There are also similar function for a specific state-action pair that called Q-function:

\begin{equation} \label{eq: Q-function}
    Q_\pi(s, a) = \mathbb{E}_{s' \sim T(.|s, a)} [r + (1-d) \cdot \gamma \cdot \mathbb{E}_{a' \sim \pi(.|f_O(s'))} Q_{\pi}(s', a')]
\end{equation}

Usually, reinforcement learning algorithms solve this problem in the assumption that environment dynamics ($T, R, f_O$ and $D$) are unknown and therefore agents can gain experience only from interactions with an environment. Such algorithms are called model-free algorithms because that they do not depend on information about environment dynamics.

One of the classic modern RL algorithms is the Deep Deterministic Policy Gradient algorithm~\cite{ddpg}. It utilizes actor-critic architecture. Critic $C$ is a neural network that estimates Q-function (eq.~\ref{eq: Q-function}) of a specific observation-action pair. It is trained by gradient descent with a loss function built upon time-difference error (TD-error): $L = \mathbb{E}_{(o, a, o', r, d} (C(o, a) - (r + \gamma \cdot (1-d) \cdot Q_{A}(o', A(o')))^2$. In practice, $Q_{A}$ is usually replaced by value yield by a target critic network that slowly updates to match the critic network. Actor $A$ is a neural network that maps observations into an action space. It is trained by a stochastic gradient descent algorithm with a loss function defined as $L = - \mathbb{E}_{o} C(o, A(o))$ in order to maximize an expectation of Q-function produced by the critic.

Another important algorithm is the Proximal Policy Optimization method~\cite{ppo}. As in DDPG, it also utilizes actor-critic architecture to solve the task with continuous observation and action spaces. In this algorithm, critic estimates value function (eq.~\ref{eq: value-function}) instead of Q-function. In PPO, the actor produces not a single action but a distribution of actions and trains to maximize the probability of actions that lead to a better outcome. 

We can interpret the Soft Actor-Critic algorithm~\cite{sac} as a combination of PPO and DDPG algorithms. Like the PPO algorithm, it uses stochastic policy instead of a deterministic one, but unlike PPO trains it to maximize Q-function directly as in DDPG. Moreover, the authors add additional action distribution entropy loss to an actor to prevent the policy from converging to a single-value distribution seen as a deterministic policy.

\subsection{Model-based Reinforcement Learning}
A model-based reinforcement learning assumes that agents have access to a world model that can predict full transitions $(s, a, s', r, d)$ by state-action pair. 

For most environments, the true world model is unknown as transition function, episode termination function or reward function are too complex or hidden from us. To deal with it, modern model-based methods~\cite{planet, dreamer, dreamer2} use trainable world models based on neural networks.

Recently, the model-based approach demonstrated good results in solving complex RL tasks by having a better sample efficiency. Specifically, the Dreamer algorithm~\cite{dreamer} was able to solve MuJoCo tasks from pixels significantly faster and better than model-free algorithms like D4PG and A3C. 

Dreamer algorithm learns the world model with the assumption of POMDP. It builds a hidden state representation using recurrent neural networks and then uses it to generate synthetic trajectories. The algorithm uses only synthetic data to train an agent, which allows to compute gradient through it and update actor by the sum of predicted rewards. During the evaluation, the world model only encodes observation into hidden state representation and then passes it to an actor that yields an action.

\begin{figure}[h]
  \centering
  \includegraphics[width=\linewidth]{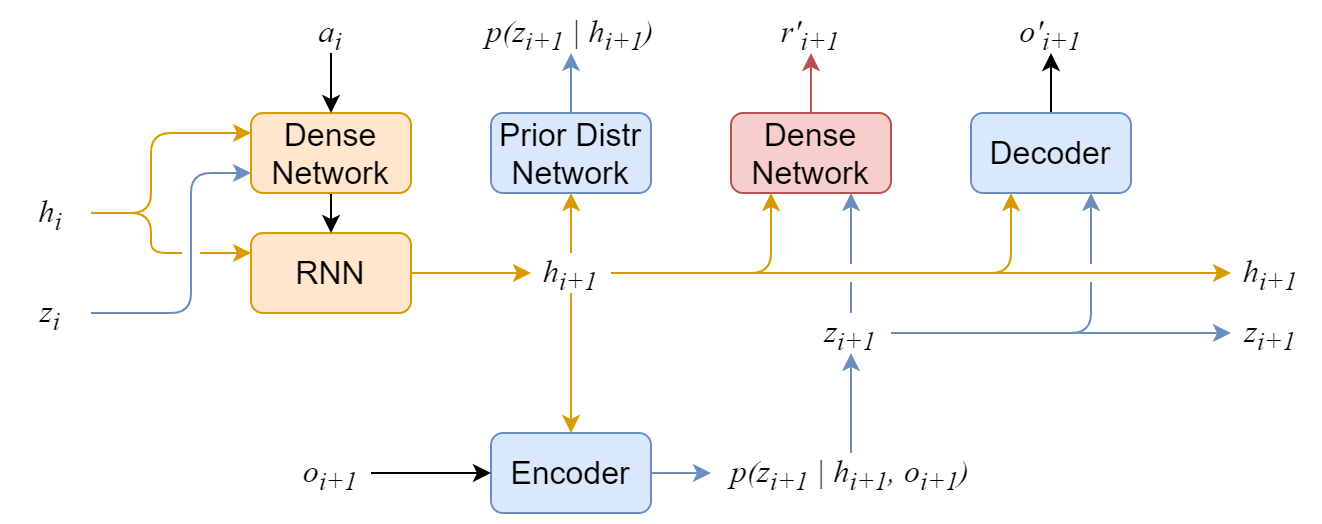}
  \caption{Original architecture of the Dreamer world model. From \cite{dreamer}.}
  \label{fig:Original Dreamer Model}
\end{figure}

The Dreamer world model (fig.~\ref{fig:Original Dreamer Model}) consists of three main parts: 
\begin{enumerate}
    \item Processing of deterministic part $h_i$ of the state (orange on fig.~\ref{fig:Original Dreamer Model}). It consists of dense and recurrent blocks. The dense block takes action and hidden state as input and passes its output to the recurrent block. Recurrent block takes output of dense block as input and $h_i$ as hidden features and outputs $h_{i+1}$.
    \item Processing of stochastic part $s_i$ of the state (blue on fig.~\ref{fig:Original Dreamer Model}). It consists of convolutional the Encoder block that estimates $p(z_i | h_i, o_i)$, and the inverse convolutional Decoder block that tries to reconstruct $o_i$ from $h_i$ and $z_i$, and a block that estimates prior distribution. $p(z_i | h_i)$. 
    \item A block that tries to predict gained reward $r_i$ from achieving state $s_i = (z_i, h_i)$.
\end{enumerate}
While interacting with environment, world model takes $h_i$, $z_i$, $a_i$ and $o_{i+1}$ as input and outputs $h_{i+1}$ and $z_{i+1} \sim p(z_{i+1} | h_{i+1}, o_{i+1})$. While generating synthetic experience, it uses prior distribution $p(z_i | h_i)$ instead of $p(z_i | h_i, o_i)$ to produce $z_i$.

To train the world model, Dreamer adapts ELBO-loss from variational autoencoders so that it encodes sequence of observations $\{o_i\}$ and decodes sequence of observations $\{o_i\}$ and rewards $\{r_i\}$. It uses $h_i$ and $z_i$ as latent representation with trainable prior distribution $p(z_i|h_i)$. The general form of loss used for training is:

\begin{equation}\label{eq: Original Dreamer Loss General}
\begin{split}
L = \mathbb{E} \sum_{i} -[&\ln p(o_t | z_i, h_i) + \ln p(r_t | h_t, z_t) \\&- \beta D_{\text{KL}}(p(z_{i+1} | z_i, h_i, o_{i+1}) \| p(z_{i+1} | h_i)]
\end{split}
\end{equation}

In practice, $p(o_t | z_i, h_i)$ and $p(r_t | h_t, z_t)$ are presented as normal distributions with fixed standard deviation. Therefore, eq.~\ref{eq: Original Dreamer Loss General} may be represented as:

\begin{equation}\label{eq: Original Dreamer Loss}
\begin{split}
L = \mathbb{E} \sum_{i} -[&\text{MSE}(o_t, o'_t) + \text{MSE}(r_t, r'_t) \\&- \beta D_{\text{KL}}(p(z_{i+1} | z_i, h_i, o_{i+1}) \| p(z_{i+1} | h_i)]
\end{split}
\end{equation}

This world model allows us to train an agent by computing gradients of cumulative discounted reward and then applying gradient ascent. Authors of the Dreamer algorithm propose to use a $\lambda$-returns estimation method to estimate cumulative discounted reward. This method was initially proposed in the Proximal Policy Optimization algorithm and works in the assumption of critic $\hat V$ that estimates value function for the current policy. $\lambda$-returns combine rollouted cumulative rewards $G_t = \sum_{i = 0}^{t} \gamma^i \cdot r_i + \gamma^{t+1} \cdot V_\pi(s_{t+1})$ as follows:

\begin{equation*}
    G_T^\lambda = (1-\lambda)\sum^{T}_{t=0} \lambda^{t} \cdot G_t + \lambda^{T} \cdot G_{0}
\end{equation*}

Which can be rewritten as:

\begin{equation} \label{eq: Lambda-returns}
    G_T^\lambda = \lambda^{T}\gamma^{T}V_\pi(s_T) + \sum_{t=0}^{T-1}\gamma^t \lambda^t \cdot (r_t + \gamma \cdot (1 - \lambda) V_\pi(s_{t+1}))
\end{equation}

Here $V_\pi(s)$ is a value function of state $s$ with respect to policy $\pi$ which in practice are replaced by its estimation $\hat V$ produced by a trained critic.

There are also several modifications of a Dreamer algorithm that improve its efficiency. For instance, the authors of Dreamer-v2~\cite{dreamer2} offers multiple changes that allow modified algorithm produce a policy that can perform as well as state-of-the-art algorithms do on the Atari Games benchmark. However, most proposed modifications are not fundamental and mostly tunes algorithms to the set of environments. Another important modification that we should mention is a Dreaming algorithm~\cite{dreaming} that replaces observation reconstruction with a contrastive learning algorithm, which makes the world model learn more informative state representations. 

\subsection{Exploration problem}
Training with no prior knowledge of POMDP dynamics forces an agent to explore new parts to find information about more optimal transitions. For some environments, finding information that will help the agent improve its policy is a complex task.  

Popular model-free algorithms often use either random actions (epsilon-greedy approach), additive random noise (as in DDPG~\cite{ddpg}), or regularization of action distribution entropy (as in SAC~\cite{sac}). However, all these methods do not aim to maximize the novelty of the agent's information explicitly. Therefore, they do not work for environments where it is necessary to commit a specific sequence of actions to reach specific parts of state space.

Plenty of exploration algorithms directly seek novel information about the environment dynamics. Some of them~\cite{neurips2020exploration1,neurips2020exploration2,curiosity,randomnetworkdistillation} motivate an agent to explore via an additional intrinsic reward. Most of these algorithms depend on a prediction error of some trainable model to measure how novel a transition is and assign an intrinsic reward depending on the novelty. 

Hazan et al.~\cite{maxentexploration} have proposed an alternative approach that maximizes the entropy of the state distribution directly, which allows an agent to cover state space completely. Proposed maximum entropy policy optimization has strong theoretical guarantees. Still, it makes some assumptions about the existence of efficient planning oracle, which makes this algorithm hardly applicable for most modern reinforcement learning tasks.

\section{Maximum Entropy Exploration in Model-based RL}
\subsection{Exploration objective}
In order to improve exploration and make it more targeted, we aim to maximize the entropy of distribution of states visited by an agent: $\max H(p_\pi(s_{\tau > t} | s_t))$. However, such distribution may have an infinite horizon which is hard to estimate. Also, as we want to improve sample efficiency by maximizing entropy of this distribution, we want our method to focus on exploring states in short horizon as they may be reached in a reasonable amount of transitions. Therefore, we will maximize an entropy of a discounted state distribution: 
\begin{equation}\label{eq: Discounted state distribution}
\begin{split}
q_\pi^{k}(s_{\tau > t+k-1} | s_t)=\ &(1 - \gamma) p_\pi^{t+k} (s_{t+k-1} | s_t) + \gamma q_\pi^{k+1} (s_{\tau > t+k} | s_t)
\end{split}
\end{equation}
where $p_\pi^{k} (s_{\tau > t} | s_t)$ is a conditional distribution of states at step $t+k$ induced by policy $\pi$. Note that this definition is suitable only for infinite episodes but can be rewritten for finite episodes by including a probability of an episode termination:
\begin{equation*}
\begin{split}
    q_\pi^{k}(s_{\tau > t+k-1} | s_t)=\ &(1 - p_\pi(d_{t+k-1} | s_{t})) \cdot [(1 - \gamma) p_\pi^{t+k} (s_{t+k-1} | s_t) + \gamma q_\pi^{k+1} (s_{\tau > t+k-1} | s_t)] \\&+ p_\pi(d_{t+k} | s_{t}) \cdot p_\pi^{t+k} (s_{t+k-1} | s_t)
\end{split}
\end{equation*} 
Where $p_\pi(d_{t+k} | s_{t})$ is a probability that episode will end after committing an action at step $t + k - 1$ conditioned to state $s_t$ at step $t$. Below we will omit termination probabilities for simplicity, but all the formulas may be easily extended to consider them. Therefore, the final objective for an agent to optimize is
\begin{equation}
\label{eq: Target Objective with Entropy}
\begin{split}
\arg\max_{\pi} (V_\pi(s_t) + \beta H(q_\pi(s_{\tau > t} | s_t))) \text{, where\ \ }q_\pi(s_{\tau > t}|s_t) := q^1_\pi(s_{\tau > t}|s_t)
\end{split}
\end{equation}

\subsection{Estimating discounted state distribution}
As distribution $q_\pi^{1}(s_{\tau > t} | s_t)$ is unknown, and we can not easily compute it from the world model, we approximate it using synthetic data generated by the world model. In model-based algorithms, the world model often generates not single synthetic transitions but transition sequences. Therefore, we may extend equation~\ref{eq: Discounted state distribution} to a sequence form:
\begin{equation}\label{eq: Discounted state distribution seq}
\begin{split}
q_\pi^{k}(s_{\tau > t+k-1} | s_t) =\ & (1 - \gamma)\sum_{i=0}^{n-1}  \gamma^i \cdot p_\pi^{t+k+i-1} (s_{t+k} | s_t) + \gamma^{n} q_\pi^{k+n} (s_{\tau > t+k+n} | s_t)
\end{split}
\end{equation}
which then can be rewritten in a recurrent form using properties of Markov Decision Process:
\begin{equation}\label{eq: Discounted state distribution reccurent}
\begin{split}
q_\pi(s_{\tau > t} | s_t)=\ &(1 - \gamma)\sum_{i=0}^{n-1} \gamma^i \cdot p_\pi^{t+i+1} (s_{t+i} | s_t) \\ & + \gamma^{n} \mathbb{E}_{s_{\tau > t+n-1} \sim p_\pi^{t+n-1} (. | s_t)} q_\pi (s_{\tau > t+n-1} | s_{t + n - 1})
\end{split}
\end{equation}

Equation~\ref{eq: Discounted state distribution reccurent} allows us to use any distribution estimation model that can sample from distribution estimation. To do so, we propose to use importance sampling in the assumption that distribution estimation $P(.|s)$ is close enough to $q_\pi(.|s)$:
\begin{equation}
\label{eq: Imprtance sampling in loss function}
\begin{split}
    \mathbb{E}_{s \sim q_\pi(. | s_t)} L(s, s_t) &= \mathbb{E}_{\{s_i \sim p_\pi^{i}(. | s_{i-1})\}_{i = t+1}^{t+n}} \mathbb{E}_{s \sim q_\pi(. | s_{t + n})} [\gamma^n L(s, s_t) + (1 - \gamma) \cdot \sum_{i=0}^{n-1}\gamma^i L(s_{i+t}, s_t)]\\
    &\approx \mathbb{E}_{\{s_i \sim p_\pi^{i}(. | s_{i-1})\}_{i = t+1}^{t+n}} \mathbb{E}_{s \sim P(. | s_{t + n})} [\gamma^n L(s, s_t) + (1 - \gamma) \cdot \sum_{i=0}^{n-1}\gamma^i L(s_{i+t}, s_t)]
\end{split}
\end{equation}

Here $L(s, s_t)$ is a loss function that is optimized by model $P$. In practice, $P$ may be any model that allows to sample from conditional distribution and estimate its density. Also, as we now have estimation $P(s|s_t)$ of distribution $q_\pi(s | s_t)$, we will replace $H(q_\pi(s | s_t))$ in equation~\ref{eq: Target Objective with Entropy} with $H(P(s|s_t))$ with the assumption that $H(q_\pi(s | s_t)) \approx H(P(s|s_t))$ if distributions are close enough. However, even for distribution estimation $P$ we may not be able to compute the entropy analytically and, moreover, not able to compute its gradient over actor network parameters. The solution is to estimate it by using importance sampling as in equation~\ref{eq: Imprtance sampling in loss function} followed by applying Monte Carlo sampling. As in SAC~\cite{sac} or DDPG~\cite{ddpg}, such estimation is theoretically consistent for any actor architecture that either deterministic or uses a reparametrization trick.

\section{Maximum Entropy Dreamer}
\label{sec: MaxEntDreamer}
\subsection{Applying Maximum Entropy Exploration}
As the Dreamer model uses learned hidden state representations in order to deal with partially observable MDP, we will maximize the entropy of the distribution of these hidden states. To estimate the discounted distribution of hidden states, we use the Mixture Density Network~\cite{mixturedensitynetwork}. It learns via the maximization of log-likelihood and estimates any distribution with a mixture of normal distributions.

In order to improve the stability of our method, we use an additional target MDN that updates every step by soft update: $\theta_T = (1 - \lambda) \theta_T + \lambda \theta$. This network is then used to estimate the entropy while updating an agent.

We also found that stochastic policy learned by the agent has a huge variance of total rewards for every iteration and mean total reward changes too much over the training time. Therefore, we add a deterministic agent that trains in parallel with a stochastic agent to make it more stable during training and evaluation time. While the goal of a stochastic agent is to explore the environment efficiently, the deterministic agent is trained to maximize the total reward.

\subsection{Additional modifications}
\label{subsec: Revised Dreamer}
\begin{figure}[h]
  \centering
  \includegraphics[width=\linewidth]{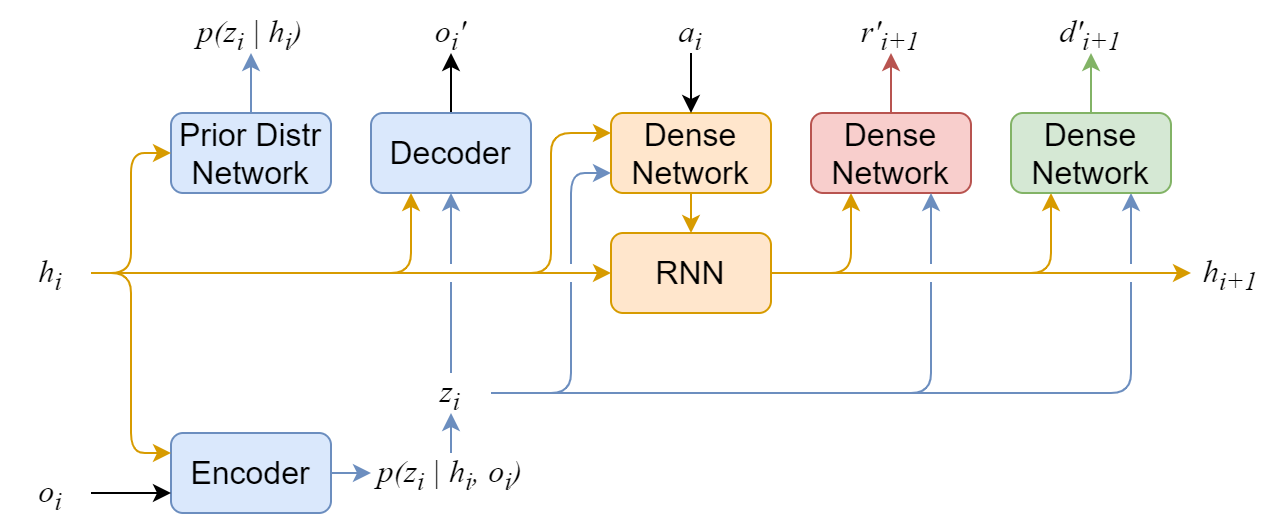}
  \caption{Architecture of the world model after all modifications. Blue part related to the stochastic part of hidden state, orange part related to the deterministic part of hidden state, red block is a reward prediction model and green block is a episode termination prediction model.}
  \label{fig:Revised Dreamer World Model}
\end{figure}

\paragraph{Episode termination prediction model.} Even though many Dreamer implementations already use such a model, we implicitly add a classification model that predicts whether a synthetic episode is finished or not. We also adjust the cumulative return formula accordingly to the presence of this model. Additionally, in the training process, we compute gradients only for the first state of synthetic sequences. This is necessary to avoid situations when the agent tries to optimize cumulative reward for states that can not be reached because of the end of the episode.

\paragraph{World model architecture rearrangement.} In the original Dreamer architecture, an agent gains information from observation only after committing its first action. Therefore, the first observations from an episode remain unobserved, leading to suboptimal performance when cumulative return depends strongly on the beginning of the episode. We rearrange the order of blocks in Dreamer's world model architecture so that agent gains information about observation before committing the first action.

\paragraph{Fixed prior distribution usage in observation encoder.} While training the world model, the original Dreamer algorithm uses trainable prior distribution $p(s_t | h_t)$ that updates along with distribution $p(s_t | h_t, o_t)$ produced by the encoder. The use of such a trainable prior will cause the stochastic part of the state vector to contain more information than intended. We fix that by adding fixed prior distribution $\mathcal{N}(0, 1)$ and optimizing $D_\text{KL}(p(s_t | h_t, o_t)\|s_t \sim \mathcal{N}(0, 1)) + D_\text{J}(p(s_t | h_t, o_t)\|p(s_t | h_t))$ instead of $D_\text{KL}(p(s_t | h_t, o_t)\|p(s_t | h_t))$. Here $D_\text{J}(P\|Q) = D_\text{KL}(P\|Q) + D_\text{KL}(Q\|P)$ is a Jeffreys Divergence. Note that we keep trainable $p(s_t | h_t)$ and use it to sample $s_t$ for an agent training.

Along with these important modifications, we also made some \textbf{minor modifications}:
\begin{enumerate}
    \item Critic now estimates Q-function instead of the Value function. This modification do not affect cumulative return estimation for agent update as $V_\pi(s) = \mathbb{E}_{a \sim \pi(.|s)} Q_\pi(s, a)$ and agent can use reparametrization trick and Monte Carlo method to estimate this mathematical expectation.
    \item For critic and reward models, we replace the MSE loss function with Smooth L1 loss function to improve training stability for higher values of reward and cumulative reward.
    \item Observation decoder now uses only stochastic part of the hidden state in order to prevent it to put too much information about observation into deterministic part of the state.
    \item While exploring, agents additionally have a probability to make a random action along with adding random noise to a greedy action.
\end{enumerate}

\section{Experiments}
\subsection{Experimental setup}
We have conducted our experiments on the set of PyBullet environments. PyBullet is an open-source physics engine which reimplements some of the popular MuJoCo environments. As the original Dreamer algorithm learns from visual information, we use rendered images of the size of $64\times64$ pixels.

As a baseline, we use the original Dreamer algorithm with hyperparameters described by its authors.

For the Revised Dreamer described in subsection~\ref{subsec: Revised Dreamer} we increase the size of the stochastic part of state representation to $128$ to compensate for the lack of information from the deterministic part of a state from the decoder. We also change the KL divergence coefficient to $0.1$ and set the same value for Jeffrey Divergence.  

Maximum Entropy Dreamer uses the same hyperparameters and coefficients as a Revised Dreamer except that it is no longer uses additive random noise for exploration. For MDN, we set the hidden size to $256$, a number of heads to $8$, the coefficient of soft update of target MDN to $0.1$, and the discount factor for discounted state distribution to $0.9$. The MDN were trained with a learning rate $0.0002$ with Adam optimizer~\cite{adam}. An entropy coefficient $\beta$ in actor optimization objective (equation~\ref{eq: Target Objective with Entropy}) is linearly decreasing from $0.2$ to $0.0001$ during the training process. 

\subsection{Experiment results}

\begin{figure}[h]
  \centering
  \includegraphics[width=\linewidth]{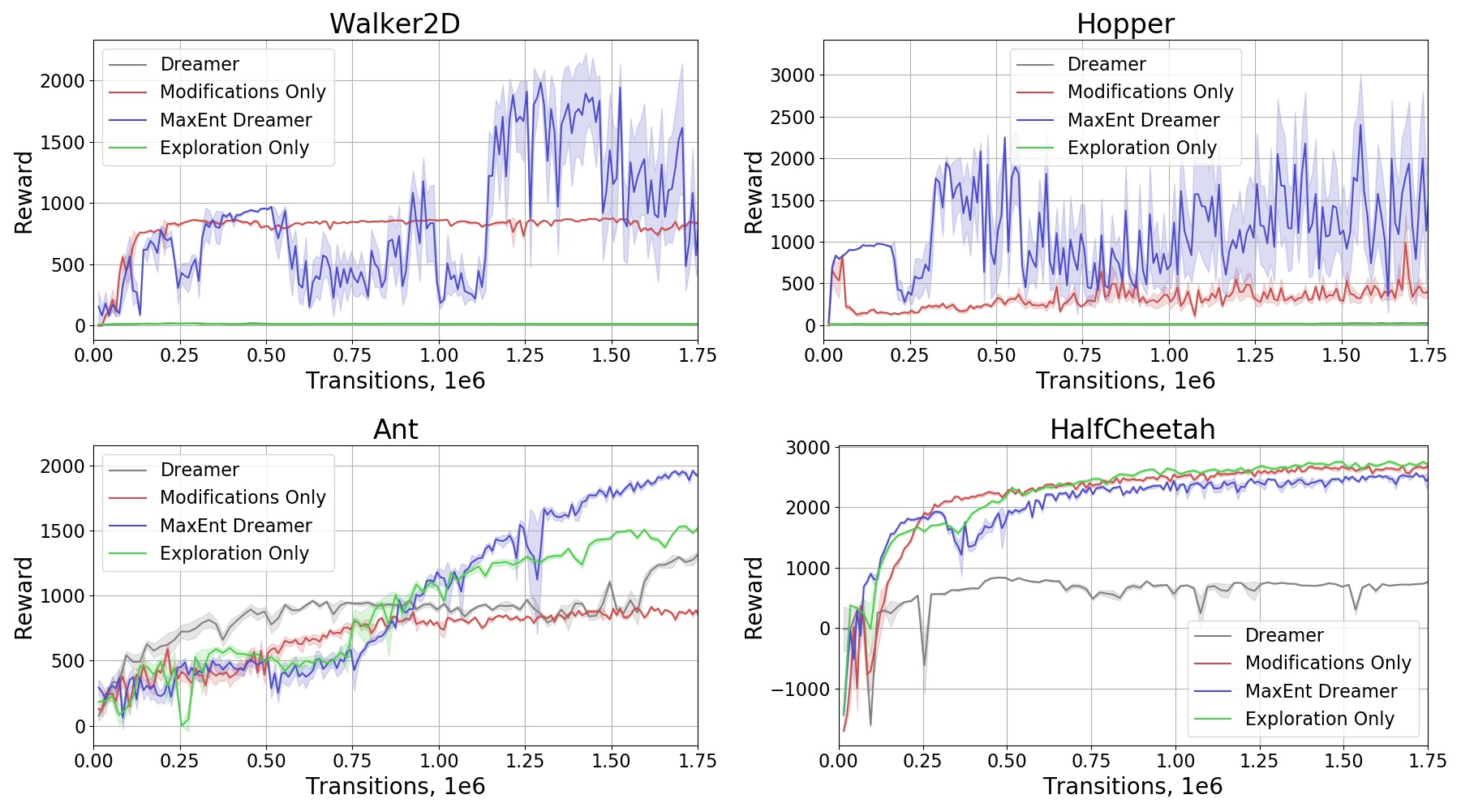}
  \caption{Comparison to baselines on PyBullet environments. Our MaxEnt Dreamer algorithm combines entropy-based exploration with some additional modifications. To evaluate the contribution of the exploration, we also show the plot for Dreamer with these additional modifications only. To measure an impact of our exploration method on original Dreamer without modifications, we show the plot for Dreamer with exploration only.}
  \label{fig:VS baselines on PyBullet}
\end{figure}

We show the results on PyBullet environments in Figure~\ref{fig:VS baselines on PyBullet}. We draw a $95\%$ confidence interval of the achieved score calculated with 10 evaluations of an agent and a line that shows a mean score.

Notably, the original Dreamer algorithm is entirely unable to solve Walker2D and Hopper environments. We believe that this is caused by the absence of a model that predicts the termination of an episode. In other words, it is too easy for an agent to meet termination conditions in these environments. Just by introducing modifications described in Section~\ref{subsec: Revised Dreamer} to address that problem, we produce an agent that is able to receive a non-zero score. Introducing entropy-based exploration improves the score even further.

In the Ant environment, aforementioned modifications do not yield any improvement, since the termination of an episode is less ambiguous. However, entropy-based exploration does significantly improves the overall performance both with and without additional modifications.

In the HalfCheetah environment, our additional modifications significantly improve the result on their own and perform on par with the full MaxEnt Dreamer and the Dreamer with our exploration method. We believe this is due to both of the reaching the optimal policy with maximal reward.

\section{Conclusion}

In this paper we present our maximum-entropy exploration method designed for model-based reinforcement learning that can be applied to any model-based RL algorithm that trains an agent as it only requires an ability to sample synthetic transitions using current policy and then compute the gradient for generated states over actor neural network parameters. 

We also employed several techniques to improve the stability of training our algorithm. The final version, MaxEnt Dreamer, shows significant improvement in 4 widely used environments. We also performed an ablation study to see what improvements are due to this stabilization alone, and demonstrate that in 3 out of 4 environments it is the exploration part that bring the majority of improvement.

As a direction for future research we suggest further investigations into the cause of instability of an agent trained with model-based maximum entropy exploration. Another direction could be to replace the Mixture Density Network that estimates discounted state distribution with more effective models.

\section{Acknowledgements}
This research was supported in part through computational resources of HPC facilities at HSE University~\cite{HPC}.

\newpage
\bibliographystyle{unsrt} 
\bibliography{main}

\begin{thebibliography}{10}

\bibitem{alphastar}
Oriol Vinyals, Igor Babuschkin, Wojciech~M Czarnecki, Micha{\"e}l Mathieu,
  Andrew Dudzik, Junyoung Chung, David~H Choi, Richard Powell, Timo Ewalds,
  Petko Georgiev, et~al.
\newblock Grandmaster level in starcraft ii using multi-agent reinforcement
  learning.
\newblock {\em Nature}, 575(7782):350--354, 2019.

\bibitem{openaifive}
Christopher Berner, Greg Brockman, Brooke Chan, Vicki Cheung, Przemys{\l}aw
  D{\k{e}}biak, Christy Dennison, David Farhi, Quirin Fischer, Shariq Hashme,
  Chris Hesse, et~al.
\newblock Dota 2 with large scale deep reinforcement learning.
\newblock {\em arXiv preprint arXiv:1912.06680}, 2019.

\bibitem{openairubicscube}
Ilge Akkaya, Marcin Andrychowicz, Maciek Chociej, Mateusz Litwin, Bob McGrew,
  Arthur Petron, Alex Paino, Matthias Plappert, Glenn Powell, Raphael Ribas,
  et~al.
\newblock Solving rubik's cube with a robot hand.
\newblock {\em arXiv preprint arXiv:1910.07113}, 2019.

\bibitem{prioritizedbuffer}
Tom Schaul, John Quan, Ioannis Antonoglou, and David Silver.
\newblock Prioritized experience replay.
\newblock {\em arXiv preprint arXiv:1511.05952}, 2015.

\bibitem{curl}
Aravind Srinivas, Michael Laskin, and Pieter Abbeel.
\newblock Curl: Contrastive unsupervised representations for reinforcement
  learning.
\newblock {\em arXiv preprint arXiv:2004.04136}, 2020.

\bibitem{drqv2}
Denis Yarats, Rob Fergus, Alessandro Lazaric, and Lerrel Pinto.
\newblock Mastering visual continuous control: Improved data-augmented
  reinforcement learning.
\newblock {\em arXiv preprint arXiv:2107.09645}, 2021.

\bibitem{spectralnormalization}
Florin Gogianu, Tudor Berariu, Mihaela Rosca, Claudia Clopath, Lucian Busoniu,
  and Razvan Pascanu.
\newblock Spectral normalisation for deep reinforcement learning: an
  optimisation perspective.
\newblock {\em arXiv preprint arXiv:2105.05246}, 2021.

\bibitem{maxentexploration}
Elad Hazan, Sham Kakade, Karan Singh, and Abby Van~Soest.
\newblock Provably efficient maximum entropy exploration.
\newblock In {\em International Conference on Machine Learning}, pages
  2681--2691. PMLR, 2019.

\bibitem{curiosity}
Deepak Pathak, Pulkit Agrawal, Alexei~A Efros, and Trevor Darrell.
\newblock Curiosity-driven exploration by self-supervised prediction.
\newblock In {\em International conference on machine learning}, pages
  2778--2787. PMLR, 2017.

\bibitem{neurips2020exploration1}
Ruo~Yu Tao, Vincent Fran{\c{c}}ois-Lavet, and Joelle Pineau.
\newblock Novelty search in representational space for sample efficient
  exploration.
\newblock {\em arXiv preprint arXiv:2009.13579}, 2020.

\bibitem{neurips2020exploration2}
Aleksandr Ermolov and Nicu Sebe.
\newblock Latent world models for intrinsically motivated exploration.
\newblock {\em arXiv preprint arXiv:2010.02302}, 2020.

\bibitem{dreamer}
Danijar Hafner, Timothy Lillicrap, Jimmy Ba, and Mohammad Norouzi.
\newblock Dream to control: Learning behaviors by latent imagination.
\newblock {\em arXiv preprint arXiv:1912.01603}, 2019.

\bibitem{dreamer2}
Danijar Hafner, Timothy Lillicrap, Mohammad Norouzi, and Jimmy Ba.
\newblock Mastering atari with discrete world models.
\newblock {\em arXiv preprint arXiv:2010.02193}, 2020.

\bibitem{ddpg}
Timothy~P Lillicrap, Jonathan~J Hunt, Alexander Pritzel, Nicolas Heess, Tom
  Erez, Yuval Tassa, David Silver, and Daan Wierstra.
\newblock Continuous control with deep reinforcement learning.
\newblock {\em arXiv preprint arXiv:1509.02971}, 2015.

\bibitem{ppo}
John Schulman, Filip Wolski, Prafulla Dhariwal, Alec Radford, and Oleg Klimov.
\newblock Proximal policy optimization algorithms.
\newblock {\em arXiv preprint arXiv:1707.06347}, 2017.

\bibitem{sac}
Tuomas Haarnoja, Aurick Zhou, Pieter Abbeel, and Sergey Levine.
\newblock Soft actor-critic: Off-policy maximum entropy deep reinforcement
  learning with a stochastic actor.
\newblock In {\em International conference on machine learning}, pages
  1861--1870. PMLR, 2018.

\bibitem{planet}
Danijar Hafner, Timothy Lillicrap, Ian Fischer, Ruben Villegas, David Ha,
  Honglak Lee, and James Davidson.
\newblock Learning latent dynamics for planning from pixels.
\newblock In {\em International Conference on Machine Learning}, pages
  2555--2565. PMLR, 2019.

\bibitem{dreaming}
Masashi Okada and Tadahiro Taniguchi.
\newblock Dreaming: Model-based reinforcement learning by latent imagination
  without reconstruction.
\newblock In {\em 2021 IEEE International Conference on Robotics and Automation
  (ICRA)}, pages 4209--4215. IEEE, 2021.

\bibitem{randomnetworkdistillation}
Yuri Burda, Harrison Edwards, Amos Storkey, and Oleg Klimov.
\newblock Exploration by random network distillation.
\newblock {\em arXiv preprint arXiv:1810.12894}, 2018.

\bibitem{mixturedensitynetwork}
Christopher~M Bishop.
\newblock Mixture density networks.
\newblock 1994.

\bibitem{adam}
Diederik~P Kingma and Jimmy Ba.
\newblock Adam: A method for stochastic optimization.
\newblock {\em arXiv preprint arXiv:1412.6980}, 2014.

\bibitem{HPC}
P.~S. Kostenetskiy, R.~A. Chulkevich, and V.~I. Kozyrev.
\newblock {HPC} resources of the higher school of economics.
\newblock {\em Journal of Physics: Conference Series}, 1740:012050, jan 2021.

\end{thebibliography}


\end{document}